\documentclass{bvm} 

\usepackage[nolist]{acronym}
\usepackage{lipsum}
\acrodef{CCMCT}[CCMCT]{canine cutanous mast cell tumor}
\acrodef{HE}[H\&E]{hematoxylin and eosin}
\acrodef{mpp}[mpp]{microns per pixel}
\acrodef{WSI}[WSI]{whole slide images}
\acrodef{MIL}[MIL]{Multiple Instance Learning}
\acrodef{ckit}[c-Kit]{c-Kit exon 11}
\begin{document}

\selectlanguage{english} 

\title{Few Shot Learning for the Classification of Confocal Laser Endomicroscopy Images of Head and Neck Tumors}

\subtitle{}

\titlerunning{Few Shot Learning for Head and Neck Tumors in CLE}

\author{
	Marc \lname{Aubreville} \inst{1},
    Zhaoya \lname{Pan} \inst{2}, 
    Matti \lname{Sievert} \inst{3},
	Jonas \lname{Ammeling} \inst{1},
    Jonathan \lname{Ganz} \inst{1},
    Nicolai \lname{Oetter} \inst{4},
    Florian \lname{Stelzle} \inst{4},
    Ann-Kathrin \lname{Frenken} \inst{5},
    Katharina \lname{Breininger} \inst{2},
	Miguel \lname{Goncalves} \inst{7}
}

\authorrunning{Aubreville et al.}

\institute{
\inst{1} Technische Hochschule Ingolstadt, Ingolstadt, Germany\\
\inst{2} Friedrich-Alexander-Universität Erlangen-Nürnberg, Erlangen, Germany\\
\inst{3} Department of Otorhinolaryngology, University Hospital Erlangen,  Friedrich-Alexander-Universität Erlangen-Nürnberg, Erlangen, Germany\\
\inst{4} Department of Oral and Maxillofacial Surgery, University Hospital Erlangen, Friedrich-Alexander-Universität Erlangen-Nürnberg, Erlangen, Germany\\
\inst{5} RWTH-Aachen University, Aachen, Germany\\
\inst{7} Department of Otorhinolaryngology, Plastic and Aesthetic Operations, University Hospital Würzburg, Würzburg, Germany}

\email{marc.aubreville@thi.de}
\maketitle

\begin{abstract}
The surgical removal of head and neck tumors requires safe margins, which are usually confirmed intraoperatively by means of frozen sections. This method is, in itself, an oversampling procedure, which has a relatively low sensitivity compared to the definitive tissue analysis on paraffin-embedded sections. Confocal laser endomicroscopy (CLE) is an in-vivo imaging technique that has shown its potential in the live optical biopsy of tissue. An automated analysis of this notoriously difficult to interpret modality would help surgeons. However, the images of CLE show a wide variability of patterns, caused both by individual factors but also, and most strongly, by the anatomical structures of the imaged tissue, making it a challenging pattern recognition task. In this work, we evaluate four popular few shot learning (FSL) methods towards their capability of generalizing to unseen anatomical domains in CLE images. We evaluate this on images of sinunasal tumors (SNT) from five patients and on images of the vocal folds (VF) from 11 patients using a cross-validation scheme. 
The best respective approach reached a median accuracy of 79.6\% on the rather homogeneous VF dataset, but only of 61.6\% for the highly diverse SNT dataset.
Our results indicate that FSL on CLE images is viable, but strongly affected by the number of patients, as well as the diversity of anatomical patterns. 
\end{abstract}

\section{Introduction}
Enhanced surgical techniques in the past decade have contributed to increased survival rates for head and neck cancer patients, with precision in tumor excision playing a pivotal role in surgical success \cite{B2}.
To achieve the fine balance between avoiding too broad excision borders and achieving a complete resection of tumor tissue, frozen sections are taken frequently during surgical removal. The frozen sections provide, however, unsatisfactory levels of sensitivity \cite{layfield2018frozen}, and extensive sampling carries the additional risk of injuring functionally important anatomical structures, as well as infections and bleeding. Confocal laser endomicroscopy (CLE) has been shown to be a non-invasive, in-vivo, in-situ real time medical imaging modality capable of providing Ear-Nose-Throat (ENT) surgeons with discriminatory features for tumor delineation~\cite{X8}. CLE enables the depiction of the superficial epithelium, usually to a predetermined depth of 60\,$\mu m$. 
Since the interpretation of these images can be complex \cite{liu2014learning}, an automatic evaluation of malignancy from CLE images could be a helpful tool for the surgeon both in the intra-operative scenario.
The classification of CLE images was shown to be feasible using neural networks \cite{aubreville2019transferability}, the task benefiting from a relatively consistent composition in terms of the malignant entity and the surface epithelium.
The classification of CLE images of the sinunasal cavity, however, carries additional challenges. For one, since CLE is not yet an established methodology for the diagnosis of ENT tumors, the number of patients that can be included in pilot studies is limited. This directly leads to a common problem in medical image recognition: Scarcity of training data for machine learning models. This problem is exacerbated by the high diversity of anatomical structures, different pathological entities and concomitant inflammatory and tumoral alterations in the sinunasal cavity, which leads to a high data diversity in CLE images. Problems like these, with high data variability and a low number of cases can be solved using few shot learning (FSL) methods, as recent success in the field of CLE demonstrates~\cite{zhou2023boosting}. FSL is a metric learning paradigm that seeks to find a latent space representation for measuring the similarity or dissimilarity of samples, enabling highly generalizable separation of classes.

In this work, we show the principal viability of FSL methods on CLE data of head and neck tumors. We assess this on multiple anatomical locations, including tumors of the oral cavity and the vocal folds, as well as sinunasal tumors. It is the first attempt at FSL from CLE images in the head and neck region.

\begin{figure}[h]
\includegraphics[width=\textwidth]{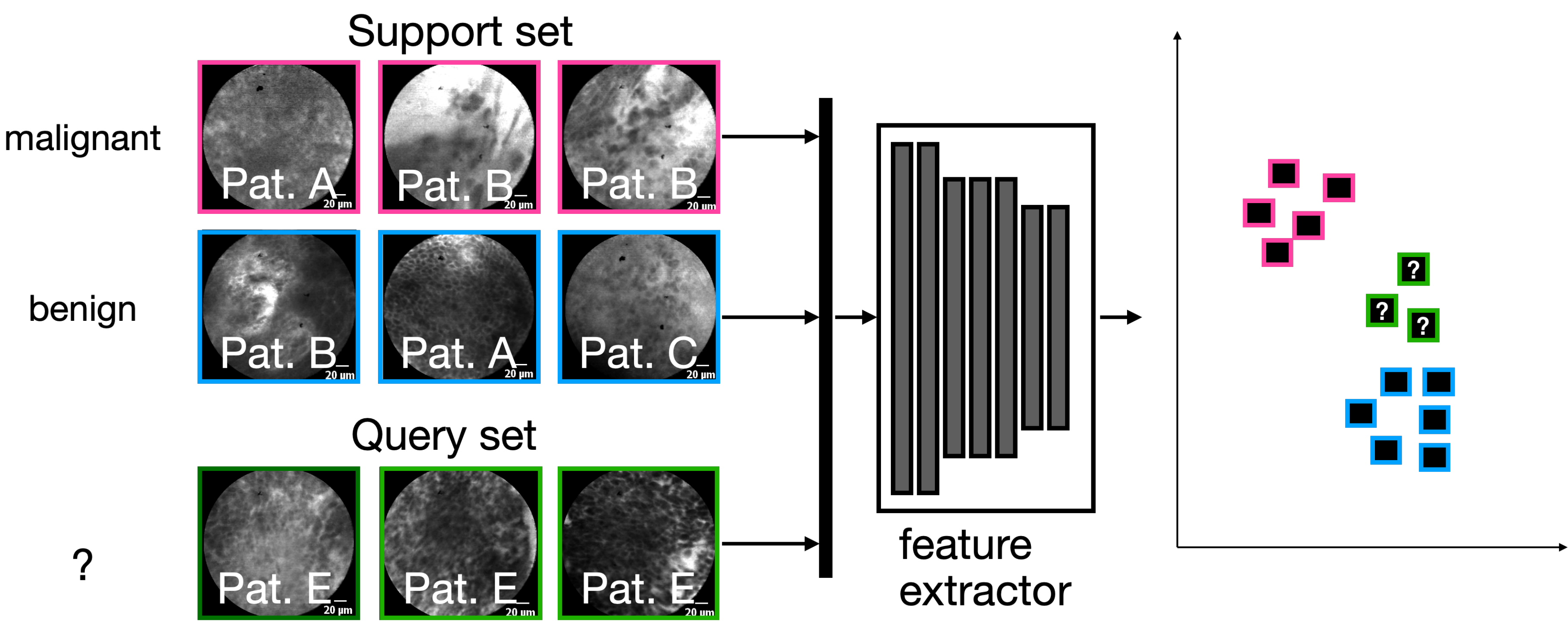}
\caption{We consider CLE malignancy classification a few shot learning task with episodic training. Patients are selected to belong to the support set or the hold out (query) set.}
\end{figure}

\section{Materials}
We used three datasets, each from a different domain of CLE images. All datasets share that they include squamous epithelium and also have a strong set of similarities in the occurring tumors. We thus assume a high potential for transfer learning between the image domains. Ethics approval was granted by the respective IRBs (243 12\ B, 60\_14 B from Universitätsklinikum Erlangen, EK 370/20 from RWTH Aachen). Frames from all CLE sequences were pre-selected to be of sufficient diagnostic quality. 

We used two previously acquired CLE datasets by our group. The first is from the oral cavity (OC) and contains twelve patients with confirmed squamous cell carcinoma (SCC) and clinically normal images from the same patients from the hard palate, inner labium, and upper alveolar ridge. The dataset contains 116 CLE sequences, totaling 7,681 frames.
The second dataset contains eleven patients and represents images acquired from the vocal folds (VF) from patients with SCC on one side, and includes images of the presumably healthy contralateral fold. The dataset contains 114 CLE sequences, totaling 7,282 frames. Prior research on this dataset suggests strong potential for transferring machine learning models trained on the OC dataset, which contains diverse anatomical conditions, to the more homogeneous VF dataset~\cite{aubreville2019transferability}. 
\begin{figure}[t]
\includegraphics[width=\textwidth]{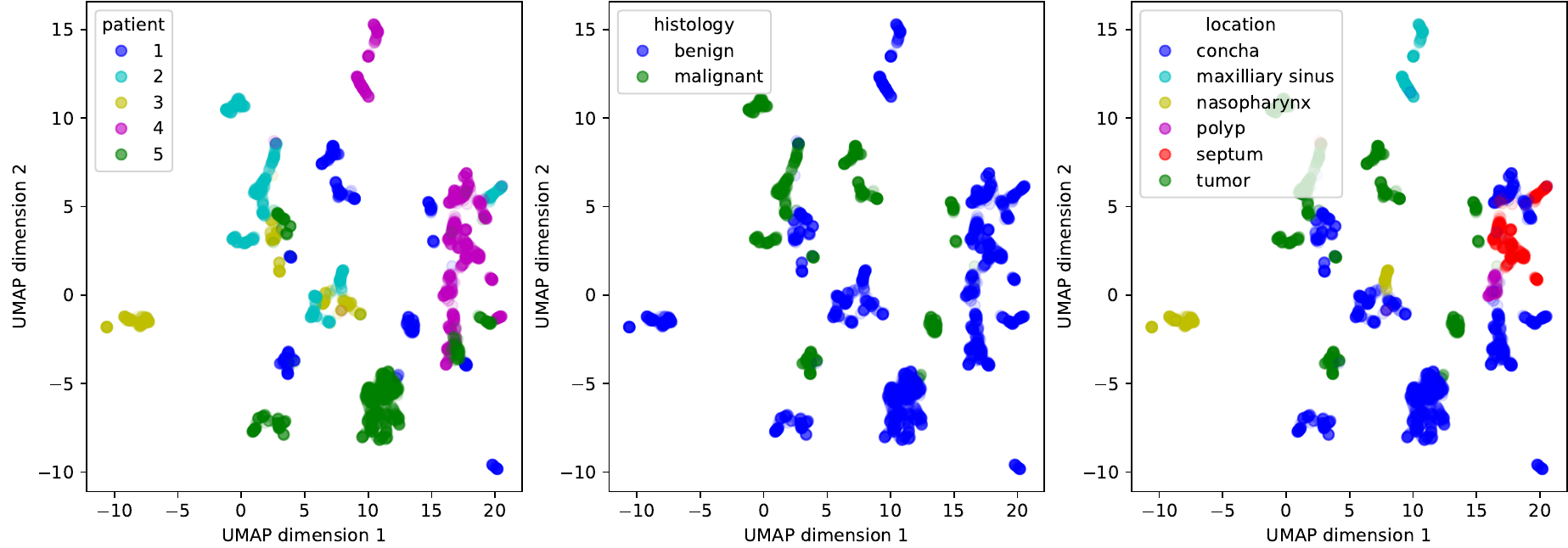}
\caption{UMAP projection of the features extracted from an ImageNet-trained ResNet18 stem of the SNT dataset. Left panel shows patients, middle panel shows malignancy status determined by histopathology, and right panel shows anatomic location.}
\label{fig:umap}
\end{figure}
Third, we used a novel dataset of sinunasal CLE images, acquired between July and September 2023 in <blinded for peer review>. 
We obtained CLE images of different locations in the nose and sinunasal cavity (septum, middle turbinate, nasopharynx, maxilliary sinus and, in one patient, a nasal polyp) as well as the area that were identified as tumorous in pre-operative radiological assessment. In total, 24 CLE sequences were recorded, each representing a different anatomical location or patient. A physician pre-selected sequences of sufficient diagnostic and image quality, resulting in 3,001 frames representing benign or non-suspicious tissues and 1,043 representing malignant carcinoma. The imaging device used as was a CellVizio system (Mauna Kea Technologies, Paris, France) with a GastroFlex probe. Fluorescein Alcon, 10\% (Alcon PHARMA, Freiburg, Germany) was used as contrast agent, of which 2.5 ml were administered prior to the investigation and 2.5 ml 8-10 minutes into the procedure to counteract degrading signal quality. Histological correlates were evaluated for all relevant anatomical positions following standard clinical procedures, yielding a malignant condition in two of the patients. An initial analysis of the dataset using an UMAP projection of a ResNet18 feature stem trained on ImageNet reveals a strong clustering in correspondence with patient and anatomical location and a less pronounced clustering with histology (Fig.~\ref{fig:umap}), highlighting the high data diversity of the task.

\section{Methods}
We compare multiple state-of-the-art FSL methods that are commonly used in similar scenarios. 
Training was always performed in 5-fold cross-validation schemes. All methods utilize a deep convolutional network as feature extractor and we used Adam as optimizer with a learning rate of $10^{-4}$. 

\subsection{Preprocessing}
CLE has a circular field of view, representing the physical cross-section of the fiber. Reflections and other effects occur at the peripheral region of this field of view and additionally, there is a strong gradient introduced by the field of view, which does not help model convergence. Therefore, we restricted the model's input to the inner rectangle of the circle. Additionally, we used standard augmentation (arbitrary rotation, flipping, blurring, sharpness adjustments) and resizing to 224$\times$224 and normalization as training image transforms.

\subsection{Few Shot Learning Methods}
FSL methods are commonly trained and evaluated using episodic training. Therein, a mini-batch that is fed to the model consists of a fixed number of support images with corresponding labels and query images that represent new samples to classify for the model and assess the potential to generalize based on these few samples from the query set. We used a five-shot setup, i.e. using five samples per class per mini-batch in the support set, and ten-fold evaluation, i.e. using ten samples per class (if available) for the query set. We used 500 episodes for training and 500 episodes for the evaluation of each method. 
Commonly, images used in FSL are assumed to be independent. However, since a high correlation of frames within patients must be expected, the split between images was carried out on the patient level, i.e. the patient of the test set was not part of the support set. 
Each of the selected FSL methods relies on having a feature extractor that was trained on a related task. For this, we used a ResNet18 feature extractor, pre-trained on ImageNet, which we  trained on the OC dataset on the same FSL task, and used the VF and SNT dataset as hold out sets of a different domain. Hypothesizing stronger generalization if the feature extractor was trained on a more diverse dataset, we also trained the feature extractor on the combined OC/VF dataset and evaluated it on the SNT dataset. In addition to the full dataset evaluation, we also provide a dataset ablation using a random subset of five patients from the VF dataset to be able to compare the properties of both datasets using similar patient counts. 

We evaluated the use of four different approaches for FSL: prototypical networks~\cite{snell2017prototypical}, the SimpleShot method \cite{wang2019simpleshot}, relation networks \cite{sung2018learning}, and matching networks \cite{vinyals2016matching}. Prototypical networks \cite{snell2017prototypical} determine a single representative (\textit{prototype}) feature vector for each class in the support set and optimize the feature extractor to minimize intra-class L2 distance and maximize inter-class L2 distance. During inference, the L2 distance of the samples of the query set are determined and the class with smallest distance is chosen. The SimpleShot method is motivated by the nearest neighbor approach and very similar, but employs a cosine distance in latent space instead of an L2 distance~\cite{wang2019simpleshot}. Relation networks learn the latent space discrimination function by using a network that combines the current feature vector with the class prototypes to provide a relation score \cite{sung2018learning}. Matching networks go one step further by attempting to learn a mapping from the extracted features (of a possibly imperfectly matching feature extractor) to a new representation where the problems are better separable given the support set \cite{vinyals2016matching}.

\begin{figure}
\includegraphics[width=\textwidth]{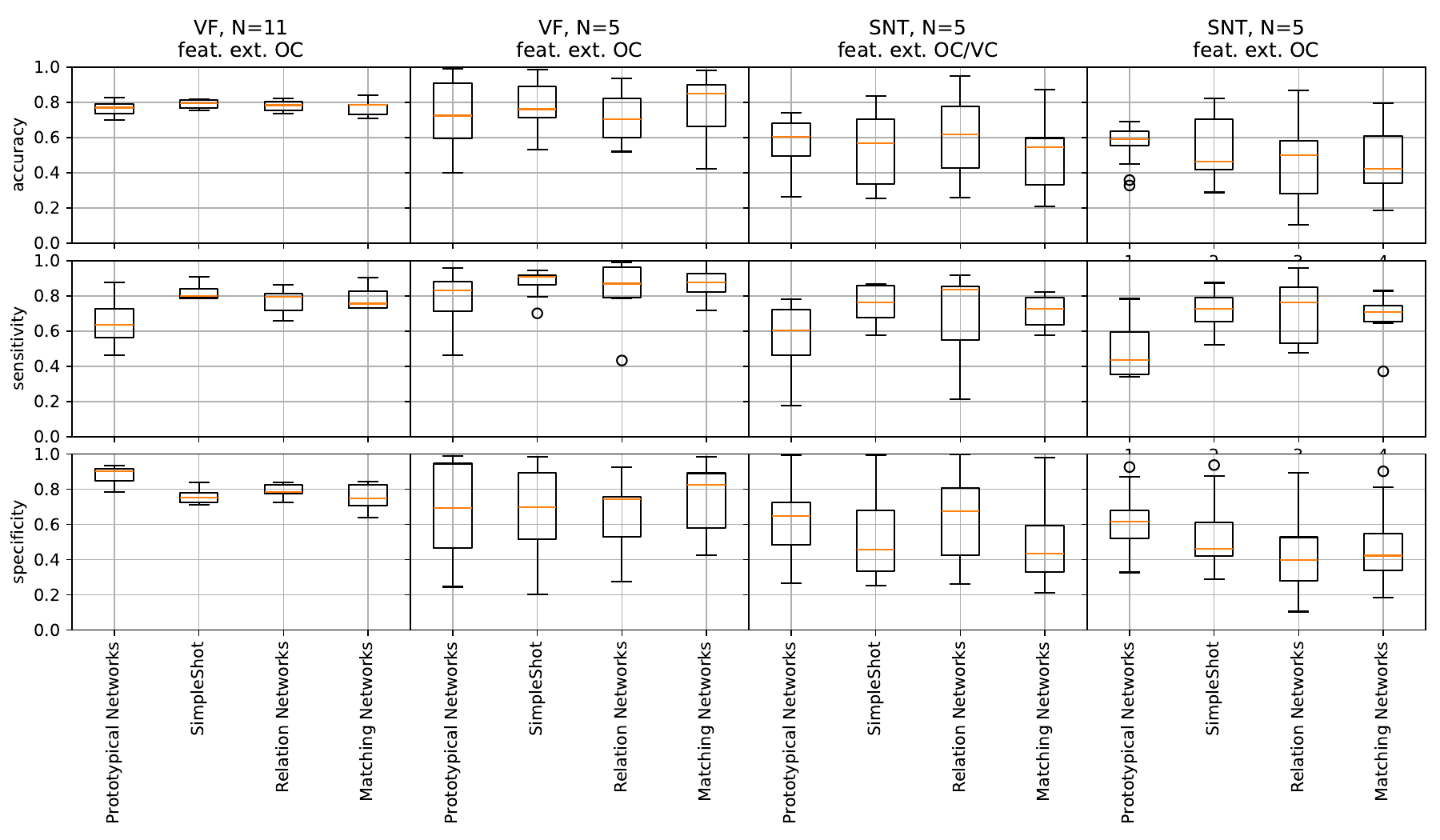}
\caption{Results of all methods on the four conditions: The full vocal folds (VF) dataset with 11 patients, an ablated random 5 patient subset thereof, and the sinunasal tumor (SNT) dataset. Training of the feature extractor was carried out on the oral cavity (OC) dataset, or, in one case on the joint OC+VF dataset.}
\label{fig:results}
\end{figure}

\section{Results}
As shown in Fig.~\ref{fig:results}, all FSL methods show similar performance on the task. We reached a median accuracy of 77.0\%, 79.6\%, 78.4\%, and 78.5\%  for prototypical networks, SimpleShot, relation networks, and matching networks on the VF dataset, respectively. When evaluated on the 5-patient subset dataset ablation, we find a median accuracy of 72.5\%, 76.1\%, 70.2\%, and 84.8\%, respectively. Overall, while all methods perform well on the full VF dataset, we find a considerable increase in result variability if the number of patients is reduced. Furthermore, the application in the SNT domain yields reduced performance across all methods (median accuracy of 59.1\%, 46.4\%, 50.0\% and 42.3\%, respectively, when trained on OC dataset), with all of them benefiting from the extended OC/VF dataset used for training the feature extractor (60.4\%, 56.9\%, 61.6\%, and 54.4\%,respectively).

\section{Discussion}
Our work confirms the findings of Zhou et al. \cite{zhou2023boosting} that FSL is, in principle, a highly capable method for the CLE imaging modality. Furthermore, we successfully demonstrated that FSL methods possess the inherent capability to achieve robust generalization across diverse anatomical domains. 
The results also indicate, however, that even though FSL methods are specifically targeted at samples with sparse data coverage, they would in fact benefit strongly from larger data sets. 
We conclude that the considerable performance drop between the evaluation on the SNT dataset and the VF dataset is likely caused by a much higher data diversity, stemming from the increased anatomical diversity found within the sinonasal cavity, as also underlined by the visual clustering of patients in Fig.~\ref{fig:umap}.
Results from human raters in our group revealed an accuracy, sensitivity and specificity of 86.9\%, 90,6\%, and 84.6\%, respectively, by human experts on a similar data set. This suggests a notable potential for further improvement with larger datasets. 

\printbibliography

\end{document}